\RequirePackage{fix-cm}
\documentclass[twocolumn]{svjour3}          
\makeatletter
\if@twocolumn
  \renewcommand\normalsize{%
   \@setfontsize\normalsize\@xpt{12.5pt}%
   \abovedisplayskip=3 mm plus6pt minus 4pt
   \belowdisplayskip=3 mm plus6pt minus 4pt
   \abovedisplayshortskip=0.0 mm plus6pt
   \belowdisplayshortskip=2 mm plus4pt minus 4pt
   \let\@listi\@listI}%

  \renewcommand\small{%
   \@setfontsize\small{8.5pt}\@xpt
   \abovedisplayskip 8.5\p@ \@plus3\p@ \@minus4\p@
   \abovedisplayshortskip \z@ \@plus2\p@
   \belowdisplayshortskip 4\p@ \@plus2\p@ \@minus2\p@
   \def\@listi{\leftmargin\leftmargini
               \parsep 0\p@ \@plus1\p@ \@minus\p@
               \topsep 4\p@ \@plus2\p@ \@minus4\p@
               \itemsep0\p@}%
   \belowdisplayskip \abovedisplayskip}

\else
  \if@smallext
   \renewcommand\normalsize{%
   \@setfontsize\normalsize\@xpt\@xiipt
   \abovedisplayskip=3 mm plus6pt minus 4pt
   \belowdisplayskip=3 mm plus6pt minus 4pt
   \abovedisplayshortskip=0.0 mm plus6pt
   \belowdisplayshortskip=2 mm plus4pt minus 4pt
   \let\@listi\@listI}%

  \renewcommand\small{%
   \@setfontsize\small\@viiipt{9.5pt}%
   \abovedisplayskip 8.5\p@ \@plus3\p@ \@minus4\p@
   \abovedisplayshortskip \z@ \@plus2\p@
   \belowdisplayshortskip 4\p@ \@plus2\p@ \@minus2\p@
   \def\@listi{\leftmargin\leftmargini
               \parsep 0\p@ \@plus1\p@ \@minus\p@
               \topsep 4\p@ \@plus2\p@ \@minus4\p@
               \itemsep0\p@}%
   \belowdisplayskip \abovedisplayskip}
 \else
  \renewcommand\normalsize{%
   \@setfontsize\normalsize{9.5pt}{11.5pt}%
   \abovedisplayskip=3 mm plus6pt minus 4pt
   \belowdisplayskip=3 mm plus6pt minus 4pt
   \abovedisplayshortskip=0.0 mm plus6pt
   \belowdisplayshortskip=2 mm plus4pt minus 4pt
   \let\@listi\@listI}%

  \renewcommand\small{%
   \@setfontsize\small\@viiipt{9.25pt}%
   \abovedisplayskip 8.5\p@ \@plus3\p@ \@minus4\p@
   \abovedisplayshortskip \z@ \@plus2\p@
   \belowdisplayshortskip 4\p@ \@plus2\p@ \@minus2\p@
   \def\@listi{\leftmargin\leftmargini
               \parsep 0\p@ \@plus1\p@ \@minus\p@
               \topsep 4\p@ \@plus2\p@ \@minus4\p@
               \itemsep0\p@}%
   \belowdisplayskip \abovedisplayskip}
  \fi
\fi
\let\footnotesize\small
\makeatother
\smartqed  
\usepackage{graphicx}
\usepackage{amsmath}
\usepackage{amssymb}

\usepackage{blindtext}
\usepackage{caption, subcaption}
\usepackage{multirow}
\usepackage{booktabs}
\usepackage{verbatim}
\usepackage{color, colortbl}
\usepackage{xspace, xcolor}
\usepackage{cite}
\usepackage[export]{adjustbox}
\usepackage{comment}
\usepackage{soul}
\usepackage{dsfont}
\definecolor{citecolor}{RGB}{34,139,34}

\usepackage{booktabs}
\usepackage{amsmath}
\usepackage{amssymb}
\usepackage{mathtools}
\usepackage{xspace}
\usepackage{hyperref}
\usepackage{graphicx}
\usepackage{caption}
\usepackage{subcaption}
\usepackage{multirow}
\usepackage{color, colortbl}
\usepackage{fancyvrb}
\usepackage{xcolor}
\usepackage{appendix}
\usepackage{natbib}

\definecolor{Gray}{gray}{0.9}

\usepackage{microtype} 
\usepackage{threeparttable}
\hypersetup{pdfstartview={FitH}}
\hypersetup{pdfborder={0 0 0}}
\hypersetup{pdfpagemode={UseNone}}
\hypersetup{colorlinks=true}
\hypersetup{citecolor=blue}
\hypersetup{linkcolor=blue}

\begin{document}

\title{Densely Distilling Cumulative Knowledge for Continual Learning}



\author{Zenglin~Shi $^{1}$       \and
        Pei Liu $^2$       \and
        Tong Su $^2$ \and
        Yunpeng Wu $^2$  \and
        Kuien Liu $^3$  \and
        Yu Song $^2$  \and
        Meng Wang $^1$
}

\authorrunning{Zenglin~Shi et al.} 

\institute{
Zenglin~Shi  \\
\email{zenglin.shi@hfut.edu.cn} \\ \\
Pei~Liu \\
\email{stcodeer@gmail.com} \\ \\
Yunpeng Wu  \\
\email{ieypwu@zzu.edu.cn} \\ \\
Meng Wang  \\
\email{wangmeng@hfut.edu.cn} \\ \\
\small $^1$ Hefei University of Technology\\
$^2$ Zhengzhou University\\
$^3$ Academy of Cyber
}

\date{Received: date / Accepted: date}

\maketitle
\def\eg{\textit{e.g.}}
\def\ie{\textit{i.e.}}
\def\Eg{\textit{E.g.}}
\def\etal{\textit{et al. }}
\def\etc{\textit{etc.}}
\newcommand{\dimny}{\mathcal{M}\xspace}
\newcommand{\dimyH}{\mathcal{H}\xspace}
\newcommand{\dimyW}{\mathcal{W}\xspace}
\newcommand{\dimH}{\ensuremath{H}}
\newcommand{\dimW}{\ensuremath{W}}
\newcommand{\dimC}{\ensuremath{C}}
\newcommand{\nStage}{\mathcal{L}\xspace}
\newcommand{\scale}{\mathcal{S}\xspace}
\newcommand{\mypartitletwo}[2][2]{\vspace*{-#1 ex}~\\{\noindent {\bf #2}}}
\newcommand{\mypartitle}[1]{\vspace*{-3ex}~\\{\noindent \underline{\bf #1}}}
\newcommand{\todo}[1]{\textcolor{red}{\textbf{#1}}}
\newcommand{\dimn}{\ensuremath{M}}
\newcommand{\apriori}{\textit{a priori}\xspace}
\newcommand{\mapping}{\ensuremath{G}\xspace}
\newcommand{\params}{\ensuremath{\theta}\xspace}
\newcommand{\data}{\ensuremath{X}\xspace}
\newcommand{\SV}{\ensuremath{X}\xspace}
\newcommand{\pro}{\ensuremath{P}\xspace}
\newcommand{\gt}{\ensuremath{G}\xspace}
\newcommand{\npro}{\ensuremath{N}\xspace}
\newcommand{\featSpace}{\ensuremath{\mathrm{\cal X}}\xspace}
\newcommand{\lSpace}{\ensuremath{\mathrm{\cal Y}}\xspace}
\newcommand{\labbb}{\ensuremath{\mathbf{t}}\xspace}
\newcommand{\state}{\ensuremath{z}\xspace}
\newcommand{\nframes}{\ensuremath{T}\xspace}
\newcommand{\kupdate}{\ensuremath{\boldsymbol{\varphi}}\xspace}
\newcommand{\sol}{\ensuremath{\boldsymbol{\beta}}\xspace}
\newcommand{\nsamples}{\ensuremath{N}\xspace}
\newcommand{\Msamp}{\ensuremath{M_{\mathrm{s}}}\xspace}
\newcommand{\nparticles}{\ensuremath{P}\xspace}

\newcommand{\nDepth}{\ensuremath{D_{\mathrm{max}}}\xspace}
\newcommand{\nTrees}{\ensuremath{K}\xspace}
\newcommand{\Xmat}{\ensuremath{\mathbf{X}}\xspace}
\newcommand{\Ymat}{\ensuremath{\mathbf{Y}}\xspace}
\newcommand{\HH}{\ensuremath{\mathbf{H}}\xspace}
\newcommand{\Smat}{\ensuremath{\mathbf{S}}\xspace}
\newcommand{\Dmat}{\ensuremath{\mathbf{D}}\xspace}
\newcommand{\eye}{\ensuremath{\mathbf{e}}\xspace}
\newcommand{\err}{\ensuremath{\boldsymbol{\xi}}\xspace}
\newcommand{\coeff}{\ensuremath{\mathbf{w}}\xspace}
\newcommand{\samp}{\ensuremath{\mathbf{x}}\xspace}
\newcommand{\laby}{\ensuremath{\mathbf{y}}\xspace}
\newcommand{\func}{\ensuremath{\mathbf{g}}\xspace}
\newcommand{\thresh}{\ensuremath{\tau}\xspace}
\newcommand{\treedepth}{\ensuremath{\Gamma_{\mathrm{depth}}}\xspace}
\newcommand{\sampler}{\emph{Sampler}}
\newcommand{\normal}{\ensuremath{\mathrm{\cal N}}\xspace}
\newcommand{\ssvmCost}{\ensuremath{\ell}\xspace}
\newcommand{\Perp}{\perp \! \! \! \perp}
\def\ci{\perp\!\!\!\perp}
\newcommand{\RR}{I\!\!R}

\newcommand{\labl}{\ensuremath{y}\xspace}

\newcommand{\mean}{\ensuremath{\mu}\xspace}

\newcommand{\Mult}{\ensuremath{\mbox{Mult}}\xspace}
\newcommand{\Cat}{\ensuremath{\mbox{Categorical}}\xspace}
\newcommand{\argmin}{\mathop{\mathrm{arg\,min}}}
\newcommand{\argmax}{\mathop{\mathrm{arg\,max}}}

\newcommand{\cs}[1]{\textcolor{red}{[\textbf{CS}: #1]}}
\newcommand{\psmm}[1]{\textcolor{orange}{[\textbf{PM}: #1]}}
\newcommand{\zl}[1]{\textcolor{blue}{[\textbf{ZL}: #1]}}
\newcommand{\sm}[1]{\textcolor{red}{[\textbf{SM}: #1]}} 

\begin{abstract}
Continual learning, involving sequential training on diverse tasks, often faces catastrophic forgetting. While knowledge distillation-based approaches exhibit notable success in preventing forgetting, we pinpoint a limitation in their ability to distill the cumulative knowledge of all the previous tasks. To remedy this, we propose Dense Knowledge Distillation (DKD). DKD uses a task pool to track the model’s capabilities. It partitions the output logits of the model into dense groups, each corresponding to a task in the task pool. It then distills all tasks’ knowledge using all groups. However, using all the groups can be computationally expensive, we also suggest random group selection in each optimization step. Moreover, we propose an adaptive weighting scheme, which balances the learning of new classes and the retention of old classes, based on the count and similarity of the classes. Our DKD outperforms recent state-of-the-art baselines across diverse benchmarks and scenarios. Empirical analysis underscores DKD's ability to enhance model stability, promote flatter minima for improved generalization, and remains robust across various memory budgets and task orders. Moreover, it seamlessly integrates with other CL methods to boost performance and proves versatile in offline scenarios like model compression.
\end{abstract}

\section{Introduction}
\label{sec:intro}
Continual learning (CL), a dynamic machine learning paradigm, focuses on models continuously adapting to new data. In contrast to traditional static training, it emphasizes a model's ability to seamlessly integrate new tasks, accumulating knowledge over time. This is crucial for maintaining the relevance and adaptability of machine learning models in evolving environments, making CL a rapidly growing field in AI. However, CL presents a dilemma: neural networks adapt rapidly to new tasks but suffer from stability issues on previous tasks. This high plasticity enables quick learning and adaptation, but low stability results in catastrophic forgetting \citep{mccloskey1989catastrophic,goodfellow2013empirical,ratcliff1990connectionist,kim2023stability}. Catastrophic forgetting leads to a significant decline in performance on older tasks after acquiring new ones.

To tackle the issue of catastrophic forgetting, weight regularization-based approaches, \eg, \citep{kirkpatrick2017overcoming,zenke2017continual,kong2023trust,aljundi2018memory,kim2023stability,chaudhry2018riemannian,aljundi2019task}, identify and preserve important parameters related to old tasks through weight regularization. Replay-based approaches, \eg, \citep{robins1995catastrophic,rebuffi2017icarl,wu2019large,hou2019learning,liu2020mnemonics,bang2021rainbow,chaudhry2018riemannian,aljundi2019gradient,isele2018selective,rolnick2019experience}, maintain a repository of exemplars from previous tasks, which are used for training alongside new tasks. Knowledge distillation-based approaches, \eg, \citep{li2017learning,castro2018end,wu2019large,mittal2021essentials,hinton2015distilling,kang2022class,douillard2020podnet,hu2021distilling,simon2021learning,peng2021hierarchical,gou2021knowledge,zhao2021soda}, leverage knowledge distillation techniques to transfer expertise from old tasks to the current model by storing previously trained models. Dynamic architecture-based approaches, \eg, \citep{hung2019compacting,yan2021dynamically,hu2023dense,yoon2017lifelong,xu2018reinforced,elsken2019neural,aljundi2017expert,wang2022foster}, construct task-specific parameters by expanding the network architecture. Each approach has its merits and drawbacks. For instance, replay-based and knowledge distillation-based approaches generally outperform weight regularization-based methods, but they require additional memory resources. Dynamic architecture-based approaches demonstrate even better performance but lead to rapidly growing model sizes as the number of tasks increases. Consequently, the choice of approach depends on the specific application context, and combining these strategies can yield enhanced forgetting prevention. 
\begin{figure}[t!]
\centering
\includegraphics[width=0.99\columnwidth]{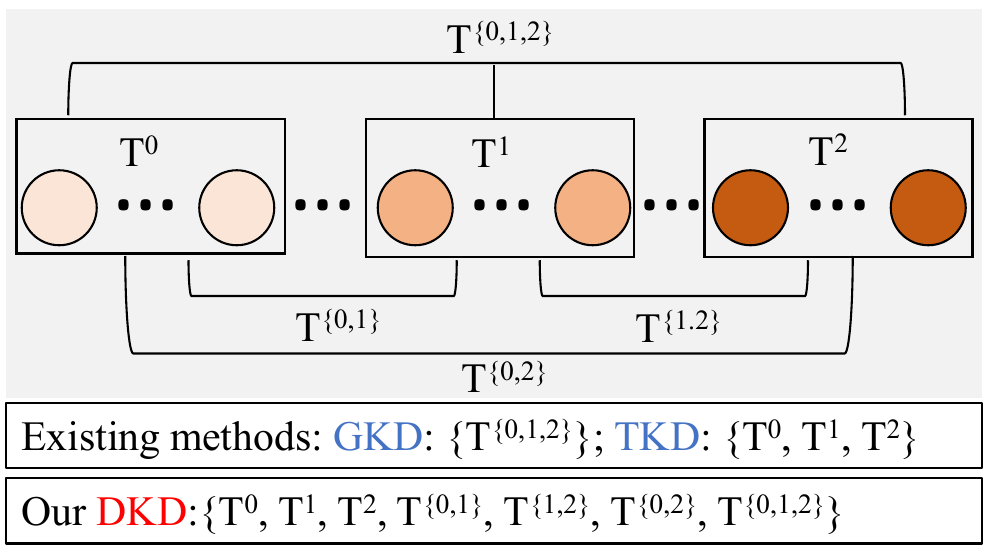}
\caption{\textbf{An illustrative example showcasing the superiority of our method.} A model that has learned $\{T^1,T^2,T^3\}$ sequentially should be able to solve them individually and jointly, as shown by the gray region. Existing distillation methods transfer knowledge for either global (\eg, GKD) or individual tasks (\eg, TKD). In contrast, our DKD transfers cumulative knowledge for all tasks and prevents forgetting better.
}
\label{fig:fig1}
\end{figure}

This work introduces a new knowledge distillation method aimed at overcoming limitations in existing approaches. Existing Global Knowledge Distillation (GKD) methods, such as \citep{rebuffi2017icarl,wu2019large,mittal2021essentials}, rely on the global softmax probability across all old tasks. On the other hand, Task-wise Knowledge Distillation (TKD) methods, such as \citep{li2017learning,castro2018end,rebuffi2017icarl}, utilize task-wise softmax probabilities obtained individually for each task. GKD transfers knowledge collectively from old tasks, while TKD transfers knowledge from old tasks individually. We propose that both global and task-specific knowledge are crucial for preventing forgetting.

In our approach, we introduce a task pool to monitor the model's capabilities across tasks. The objective is for the model, after training in each incremental step, to excel in all tasks within the task pool. However, existing GKD and TKD methods ensure effective performance only on a small subset of tasks within the pool, hindering the accumulation of knowledge across tasks, as illustrated in Fig. \ref{fig:fig1}. To tackle this, we partition the model's output logits into dense groups based on the task pool, with each group corresponding to a specific task. Distillation is then performed using all these groups. Given the computational expense of utilizing all groups, we additionally suggest uniformly and randomly selecting a task group for distillation in each optimization step. With sufficient optimization steps, all tasks in the task pool can be selected for distillation. Consequently, this random dense distillation achieves a similar effect as full dense distillation but with reduced computational complexity. To strike a balance between learning new classes and retaining knowledge of old classes, we introduce an adaptive weighting scheme based on the number and similarity of new and old classes. 

To summarize, we highlight the following contributions. 
\begin{itemize}
\item We introduce a task pool to monitor the model's capabilities across tasks. We highlight that existing GKD and TKD approaches fall short in distilling cumulative knowledge, as they only ensure performance on a subset of tasks within the pool (Section \ref{sec:pool}).  
\item We propose Dense Knowledge Distillation (DKD), distilling all cumulative knowledge by partitioning output logits into dense groups based on the tasks in the task pool. While using all groups is computationally expensive, we suggest random group selection in each optimization step. An adaptive weighting scheme based on new and old class characteristics enhances the balance between learning new classes and retaining old knowledge (Section \ref{sec:dkd}).
\item Our method outperforms recent state-of-the-art baselines across various benchmarks and scenarios. Experiment anaysis shows that it prevents forgetting by enhancing model stability, and improves generalization by promoting flatter minima. Moreover, it is robust to different memory budgets and task orders. Importantly, it seamlessly integrates with other CL methods to enhance their performance further, and demonstrates versatility in offline scenarios like model compression. (Section \ref{sec:results}).
\end{itemize} 

\section{Related work}\label{sec:relatedwork}

\subsection{Catastrophic forgetting in continual learning}
\label{sec:cl}
Continual learning has three main scenarios \citep{van2019three}. Class-incremental learning lets the model learn new classes and remember old ones. Task-incremental learning makes the model adapt to new tasks or domains, starting from one or more tasks and adding new ones as needed. Domain-incremental learning makes the model cope with changes in data or environment, learning from one domain and adapting to new ones without losing its original domain. Catastrophic forgetting is a major challenge in continual learning. Researchers have proposed many approaches to address this issue. Weight regularization-based approaches, \eg, \citep{kirkpatrick2017overcoming,zenke2017continual,aljundi2018memory,lee2020continual}, aim to enforce stringent constraints on model parameters by penalizing changes in parameters crucial for old tasks. Knowledge distillation-based approaches, \eg, \citep{li2017learning,lee2019overcoming,hou2018lifelong,zhang2020class,castro2018end,wu2019large,douillard2020podnet,mittal2021essentials}, typically involve storing a snapshot of the model learned from old tasks in a memory buffer and then distilling the knowledge encoded in the stored model to the current model. Replay-based approaches, \eg, \citep{robins1995catastrophic,rebuffi2017icarl,castro2018end,wu2019large,hou2019learning,liu2020mnemonics} store a few exemplars from old tasks in a memory buffer, and many of them use herding heuristics \citep{welling2009herding} for exemplar selection. The stored exemplars combined with the new data are used to optimize the network parameters when learning a new task. Architecture-based approaches, \eg, \citep{aljundi2017expert,rajasegaran2019random,hung2019compacting,yan2021dynamically,hu2023dense,wang2022foster,zhou2022model,schwarz2018progress,pham2021dualnet,zhao2021mgsvf}, establish task-specific parameters explicitly to address this challenge. 

In this work, we propose a novel knowledge distillation method to address the issue of catastrophic forgetting. Next, we present and discuss several existing knowledge distillation-based techniques closely linked to our work, highlighting the unique contributions and innovations we bring to this field.

\subsection{Knowledge distillation-based approaches}
To facilitate knowledge transfer from an old model to a new one and mitigate forgetting, knowledge distillation (KD), initially introduced by \cite{hinton2015distilling}, offers an intuitive approach. KD enables the teaching of the new model using the insights of a teacher model. Establishing this distillation relationship involves various techniques. For example, LwF \citep{li2017learning} was among the first to successfully apply knowledge distillation to continual learning. It does so by introducing a regularization term via knowledge distillation to combat forgetting. This term enforces alignment between old and new models by making predicted probabilities of old classes the same. Since the predicted probabilities of old classes are computed for distillation in a task-wise manner, this kind of distillation is called Task-wise KD (TKD), which is widely used by subsequent works such as EEIL \citep{castro2018end} and iCaRL \citep{rebuffi2017icarl}. Different from TKD, Global KD (GKD) computes the global predicted probability across all old classes, which is also widely used in subsequent works such as BIC \citep{wu2019large} and CCIL \citep{mittal2021essentials}.

In addition to distilling logits, some approaches focus on distilling intermediate features of deep models. For example, UCIR \citep{hou2019learning} encourages the orientation of features extracted by the new model to be similar to those by the old model via distillation. PODNet \citep{douillard2020podnet} minimizes the differences in pooled intermediate features along height and width directions instead of performing element-wise distillations. AFC \citep{kang2022class} conducts distillation while considering the importance of various feature maps. To reveal structural information in model distillation, several works also suggest conducting relational knowledge distillation \citep{park2019relational}. For example, R-DFCIL \citep{gao2022r} employs structural inputs (e.g., triples) and aligns the input relationship of the old and new model.

Our proposed method falls under the category of logit distillation but introduces a novel aspect by incorporating a task pool to monitor the model's capabilities across tasks. Unlike existing GKD and TKD approaches, which focus on a subset of tasks, our method distills all cumulative knowledge by partitioning output logits into dense task-wise groups. Importantly, our approach is compatible with feature and relational distillation methods, enabling seamless integration and potential enhancements when combined.

\section{Method}
\subsection{Problem formulation.} 
This paper focuses on the class-incremental scenario in CL, where the objective is to learn a unified classifier over incrementally occurring sets of classes \citep{rebuffi2017icarl}. 
Formally, let $\mathcal{T} = \{\mathcal{T}^0, \ldots, \mathcal{T}^T\}$ represent a set of classification tasks involving classes $\{C^0, \ldots, C^T\}$. A single model $f$ is trained to solve the tasks $\mathcal{T}$ incrementally. In each incremental step, the parameters of $f$ are fine-tuned to align with the learning objectives of the new classes. Let $\mathcal{L}^{t-1}$ and $\mathcal{L}^t$ denote the learning objectives for tasks $\mathcal{T}^{t-1}$ and $\mathcal{T}^t$, respectively. Additionally, consider $w^{t-1}$ and $w^t$ represent the convergent or optimal parameters after sequential training on tasks $\mathcal{T}^{t-1}$ and $\mathcal{T}^t$, respectively. As $\mathcal{L}^{t-1}$ and $\mathcal{L}^t$ are mutually independent, a notable disparity exists between $w^{t-1}$ and $w^t$. Consequently, the model $f$ with parameters $w^t$, struggles to perform well on task $\mathcal{T}^{t-1}$ while excelling on task $\mathcal{T}^t$, leading to the challenge of catastrophic forgetting.

\textbf{Knowledge distillation.}
To address the challenge of forgetting, the learning process for a new task incorporates a knowledge distillation (KD) regularized loss $\mathcal{L} = \mathcal{L}_{CLS}+\lambda \mathcal{L}_{KD}$. Here, the classification loss  $\mathcal{L}_{CLS}$ is employed for learning the new classes, while the distillation loss $\mathcal{L}_{KD}$ serves as a regularization term to retain the knowledge of the old classes through KD. The trade-off term $\lambda$ is used to balance the learning of new classes and the retention of knowledge from old classes.

Consider a (image, label) pair $(x^{t},y^{t})$ drawn from the training set of the task $\mathcal{T}^t$. Let $z^{t}=f^{t}(x^{t})$ represent the output logit of the new model $f^{t}$ and $z^{t-1}=f^{t-1}(x^{t})$ the output logit of the old model $f^{t-1}$. The classification loss $\mathcal{L}_{CLS}=\ell_{CE}(\sigma(z^{t}_{C^t}),y^t)$  is defined with $\sigma(\cdot)$ as the softmax function, $z^{t}_{C^t}$ indicating the output logits corresponding to the new classes $C^t$, and $\ell_{CE}$ representing the standard cross-entropy loss. As previously discussed, two versions of $\mathcal{L}_{KD}$, namely Global KD (GKD) and Task-wise KD (TKD), are defined for learning task $t$ as follows: 
\begin{equation}
\mathcal{L}^t_{GKD} = \ell_{KL}\big(\sigma(z^{t}_{C^0:C^{t-1}}/\tau), \sigma(z^{t-1}_{C^0:C^{t-1}}/\tau)\big).
\label{eq:gkd}
\end{equation}
\begin{equation}
\mathcal{L}^t_{TKD} = \sum_{i=1}^{t-1}\ell_{KL}\big(\sigma(z^{t}_{C^i}/\tau), \sigma(z^{t-1}_{C^i}/\tau)\big).
\label{eq:tkd}
\end{equation}
Here, $\ell_{KL}$ denotes the standard Kullback-Leibler (KL) divergence. The temperature hyper-parameter $\tau$ is conventionally set to $2$. $z^{t}_{C^0:C^{t-1}}$ and $z^{t-1}_{C^0:C^{t-1}}$ refer to the output logits for all old classes $\{C^0, \ldots, C^{T-1}\}$. Similarly, $z^{t}_{C^i}$ and $z^{t-1}_{C^i}$ denotes the output logits for the old classes learned in the task $i$. 

\textbf{Our Motivation.} In Eq. (\ref{eq:gkd}), GKD utilizes the global softmax probability across all old classes, denoted as $\sigma(z_{C^0:C^{t-1}}/\tau)$. In contrast, TKD in Eq. (\ref{eq:tkd}) employs task-wise softmax probabilities, denoted as $\{\sigma(z_{C^i}/\tau\}_{i=1}^{t-1}$, obtained separately for each task. GKD transfers knowledge collectively from the old classes, while TKD transfers knowledge from the old classes individually. We posit that both global knowledge and task-specific knowledge should be transferred from the old model to the new model to prevent forgetting. To achieve a more nuanced distillation of cumulative knowledge within the old model, we introduce a new approach termed Dense KD (DKD).

\begin{figure*}[t!]
\centering
\includegraphics[width=1.99\columnwidth]{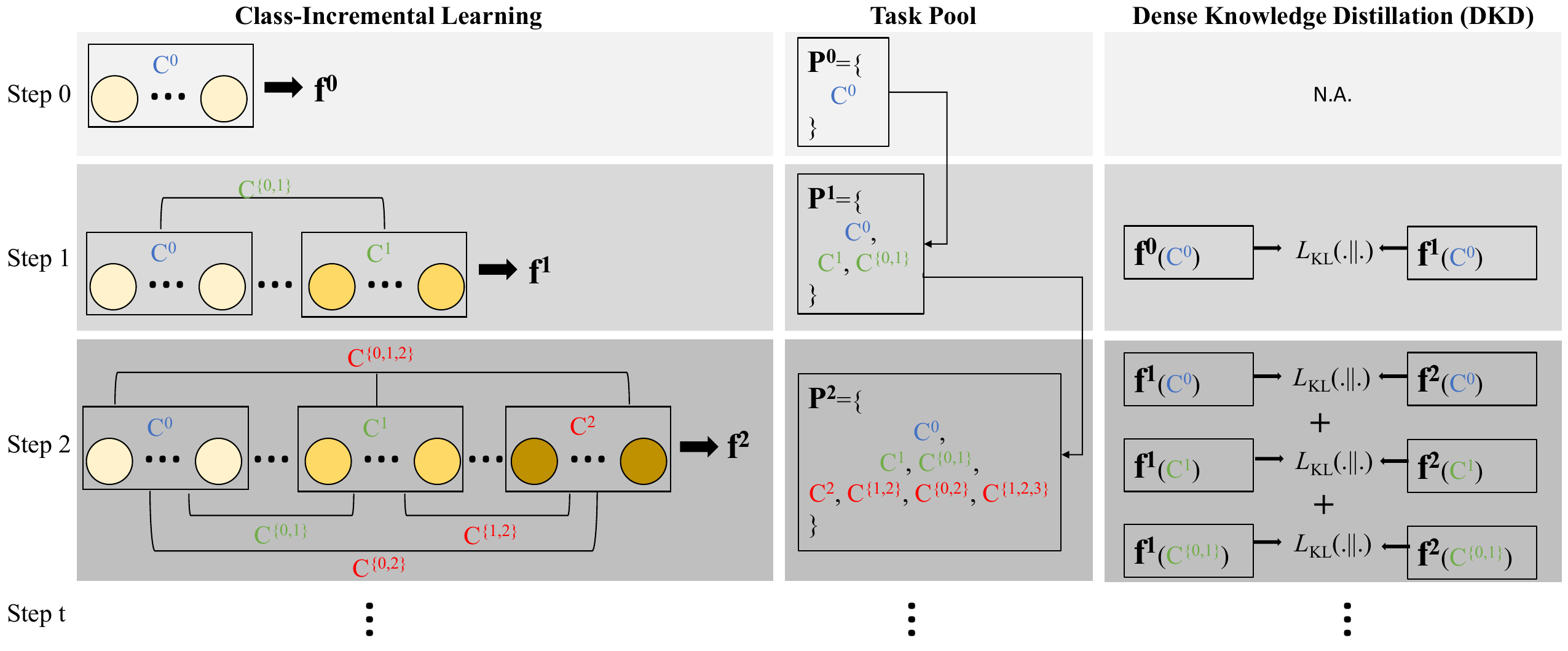}
\caption{\textbf{Illustration of our approaches.} When incrementally learning three classification tasks involving classes $\{C^0, C^1, C^2\}$ using a model $f$, a task pool $P$ monitors the tasks within the model's capabilities. The ideal capability of model $f$ includes recognizing task-specific classes individually and their combined classes. Our DKD facilitates the transfer of cumulative knowledge for recognizing both task-specific and combined classes, as indicated in the task pool, from earlier models to the new model. 
}
\label{fig:rdkd}
\end{figure*}

\subsection{Task pool}
\label{sec:pool}
Before delving into our approach, let's introduce the concept of a task pool, denoted as $P$, proposed to monitor the tasks within the capabilities of the model $f$, as illustrated in Fig. \ref{fig:rdkd}. In the initial step, the model $f$ undergoes training to tackle a classification task involving classes $C^0$, resulting in the model $f^0$ and the task pool $P^0 = \{C^0\}$. In the first incremental step, $|C^1|$ new output nodes are added to the output layer of $f$ to accommodate additional classes $C^1$. The classification loss $\mathcal{L}_{CLS}$ ensures $f^1$ recognizes classes $C^1$. The knowledge of $C^0$ is transferred from $f^0$ to $f^1$ by using either GKD loss $\mathcal{L}_{GKD}$ or TKD loss $\mathcal{L}_{TKD}$. Consequently, classification loss and distillation loss together ensure $f^1$ recognizes both classes $C^0$ and $C^1$ simultaneously. As a result, the task pool is updated to $P^1 = P^0 \uplus C^1=\{C^0,C^1, C^{\{0,1\}}\}$, where $\uplus$ denotes the merging and combining operation, and $C^{\{0,1\}}$ represents the task of recognizing both classes $C^0$ and $C^1$ simultaneously.

In the second incremental step, an additional $|C^2|$ output nodes are added to the output layer to accommodate new classes $C^2$. The classification loss $\mathcal{L}_{CLS}$ ensures that $f^2$ recognizes the new classes $C^2$. However, either the GKD loss $\mathcal{L}_{GKD}$ or the TKD loss $\mathcal{L}_{TKD}$ can only transfer part of the knowledge from $f^1$ to $f^2$. Specifically, $\mathcal{L}_{GKD}$ ensures that $f^2$ acquires the knowledge of $f^1$ on the task of recognizing $C^{\{0,1\}}$, overlooking its knowledge on the specific task of recognizing $C^0$ and $C^1$ separately. Consequently, $f^2$ may perform well only on tasks of $\{C^2, C^{\{0,1\}}\}$ by using $\mathcal{L}_{GKD}$. In contrast, $\mathcal{L}_{TKD}$ ensures that $f^2$ acquires the knowledge of $f^1$ on the task of recognizing $C^0$ and $C^1$ separately, overlooking its knowledge on the task of recognizing $C^{\{0,1\}}$. Consequently, $f^2$ may perform well only on tasks of $\{C^0, C^1, C^2\}$ by using $\mathcal{L}_{TKD}$. However, the ideal capability of the model $f^2$ should encompass the recognition of classes $C^0$, $C^1$, and $C^2$ individually, as well as their combined classes. Therefore, the task pool should be updated to $P^2 = P^1 \uplus {C^2}=\{C^0, C^1, C^2, C^{\{0,1\}}, C^{\{0,2\}}, C^{\{1,2\}}, C^{\{0,1,2\}}\}$.

This iterative update of the task pool continues in subsequent steps, with the task pool in the $t$-th incremental step denoted as $P^t = P^{t-1} \uplus {C^t}$, and so forth. However, GKD in Eq. (\ref{eq:gkd}) and TKD in Eq. (\ref{eq:tkd}) only ensure that the model $f$ performs well on a small subset of tasks in the task pool $P$. This limitation hampers the accumulation of knowledge across tasks and weakens model stability, as supported by empirical evidence (see \textbf{Section} \ref{sec:analysis}).

\subsection{Dense knowledge distillation}
\label{sec:dkd}
\textbf{Full dense knowledge distillation (FDKD).} To effectively transfer all cumulative knowledge from earlier models to the new model for enhancing model stability, we propose to conduct dense distillations instead of relying on the single global distillation outlined in Eq. (\ref{eq:gkd}) or the simple task-wise distillation outlined in Eq. (\ref{eq:tkd}). Specifically, we categorize the output logits of the old model $f^{t-1}$ into dense groups based on the task pool $P$, where each group corresponds to a task within the task pool. Subsequently, we perform dense distillations using all groups, enabling the comprehensive transfer of accumulative knowledge from the old model $f^{t-1}$ to the new model $f^{t}$, as illustrated in Fig. \ref{fig:rdkd}. Then we can obtain a new distillation regularized loss:
\begin{equation}
\mathcal{L}_{FDKD}^t = \sum_{p \in P^{t-1}}\ell_{KL}\big(\sigma(z^{t}_p/\tau), \sigma(z^{t-1}_p/\tau)\big).
\label{eq:fdkd}
\end{equation}
Here, $z^{t}_p$ and $z^{t-1}_p$ denotes the grouped output logits for the old classes $p$ from task pool $P^{t-1}$. 

\textbf{Random dense knowledge distillation (RDKD).} Eq. (\ref{eq:fdkd}) ensures the cumulative knowledge present in the old model $f^{t-1}$ is entirely distilled to the new model $f^t$. However, using all the task groups for distillation in every optimization step is very computationally expensive. Thus, we also suggest random group selection for distillation. Then we rewrite Eq. (\ref{eq:fdkd}) to,
\begin{equation}
\mathcal{L}_{RDKD}^t = \ell_{KL}\big(\sigma(z^{t}_p/\tau), \sigma(z^{t-1}_p/\tau)\big), p \in P^{t-1}.
\label{eq:rdkd}
\end{equation}
Here, $p$ is uniformly and randomly selected from the task pool $P$ in each optimization step (iteration). Given that the model $f^t$ is usually well-trained with sufficient optimization steps, all tasks in the task pool $P$ can be selected for distillation. Consequently, this random dense distillation exhibits a similar effect as the full dense distillation defined in Eq. (\ref{eq:fdkd}) and carries the same computational complexity as the GKD outlined in Eq. (\ref{eq:gkd}), as supported by empirical evidence (see section \ref{sec:analysis}).

\textbf{Adaptive weighting scheme.} We additionally introduce an adaptive weighting scheme to determine an appropriate value for $\lambda$. This scheme comprises two components: the ratio of the count of old classes in task $P$ to the count of new classes and the similarity between old classes and new classes:
\begin{equation}
\lambda^t = \lambda_{base}*r(|p|,|C^t|)*s(p,C^t).
\label{eq:rkd}
\end{equation}
Here, $\lambda_{base}$ is a fixed constant for each dataset. $r(|p|,|C^t|) = \sqrt{|p|/|C^t|}$ denotes the ratio, and $|p|$ and $|C^t|$ are the count of old classes in the task $P$ and the count of new classes in the $t$-th incremental step. $s(p,C^t)$ denotes the similarity. $r(|p|,|C^t|)$ strikes a balance between learning new classes and remembering old classes based on their class count. To compute the similarity, we compute a feature vector for each class after the learning of each task, \ie, $v = \frac{1}{n}\sum_{i=0}^{n} GAP\big(f_{enc}(x_i)\big)$ where there are $n$ training samples $x$ for each class, $f_{enc}$ is the feature extractor encoder of the model $f$, and GAP denotes global average pooling. The dimensionality of the vector $v$ corresponds to the count of channels in the output feature maps of $f_{enc}$. Then we compute the similarity between old classes in task $p$ and new classes $C^t$ by measuring their Euclidean distance, \ie, $s(p,C^t)=\sqrt{\sum_{j=0}^{m} (\Bar{v}_j^{p}-\Bar{v}_j^{C^t})^2}$ where $\Bar{v}_j^{p}$ is the mean feature vector of all classes in the task $p$, and $\Bar{v}_j^{C^t}$ is the mean feature vector of all new classes $C^t$. Smaller weight values are assigned if new and old classes are more similar, while larger weight values are assigned if they are less similar. This adaptive weighting allows a focus on preserving the knowledge of old classes that are more susceptible to forgetting.
\begin{table*}[t!]
\centering
\footnotesize
\caption{\textbf{The effect of the proposed RDKD on CIFAR100}. RDKD shows the highest average incremental accuracy among the compared approaches across four different protocols. Also, RDKD seamlessly integrates with existing methods, resulting in a significant enhancement in accuracy across all protocols. The improved performance over existing methods is indicated by the numbers highlighted in \textcolor{blue}{blue}.}
\resizebox{1.8\columnwidth}{!}{
\begin{tabular}{@{}lccccccccccccccccccccc@{}}
\toprule
&\multicolumn{4}{c}{\textbf{CIFAR100}} \\
\cmidrule(lr){2-5}
&\textbf{T=5} & \textbf{T=10} & \textbf{T=25} & \textbf{T=50}\\
\hline

BIC \citep{wu2019large} &56.82 &53.21 &48.96 &47.09 \\
iCaRL \citep{rebuffi2017icarl}&58.16 &53.57 &51.06 &44.35 \\
UCIR \citep{hou2019learning} &63.24 &61.22 &57.57 &49.30 \\
Podnet \citep{douillard2020podnet} &64.20 &62.11 &60.82 &57.13 \\
CCIL \citep{mittal2021essentials} &65.73 &64.40 &60.67 &56.27 \\
AFC \citep{kang2022class} &66.49 &64.41 &63.24 &61.53 \\
\rowcolor{Gray}
RDKD (\textit{ours}) &\textbf{67.51} &65.06 &62.31 &57.26 \\
\midrule
iCaRL+RDKD (\textit{ours}) &59.95 (\textcolor{blue}{+1.83}) &56.98 (\textcolor{blue}{+3.45}) &52.56(\textcolor{blue}{+1.52}) &45.64(\textcolor{blue}{+1.29}) \\
Podnet+RDKD (\textit{ours}) &65.95 (\textcolor{blue}{+1.75}) &62.87 (\textcolor{blue}{+0.7}) &61.17(\textcolor{blue}{+0.89}) &58.12(\textcolor{blue}{+0.99}) \\
CCIL+RDKD (\textit{ours}) &67.43 (\textcolor{blue}{+1.70}) &64.73 (\textcolor{blue}{+0.33}) &61.17(\textcolor{blue}{+0.5}) &56.46(\textcolor{blue}{+0.19}) \\
AFC+RDKD (\textit{ours}) &67.13 (\textcolor{blue}{+1.13}) &\textbf{65.35} (\textcolor{blue}{+0.95}) &\textbf{63.81}(\textcolor{blue}{+0.57}) &\textbf{62.17}(\textcolor{blue}{+0.64}) \\
\bottomrule
\end{tabular}}
\label{tab:cifar100}
\end{table*}
\section{Experiments and results}
\label{sec:results}
\subsection{Experimental setup}
\label{sec:setup}
\textbf{Datasets.}
We conduct experiments on three standard datasets: CIFAR100, ImageNet100, and ImageNet1000, following previous works, \eg, \citep{hou2019learning,liu2020mnemonics,tao2020topology,douillard2022dytox,russakovsky2015imagenet}. 
CIFAR100 consists of $60,000$ color images with a fixed size of $32 \times 32$ pixels, divided into 100 classes.
The dataset has $50,000$ training images and $10,000$ testing images. 
ImageNet1000 contains about $1.3$ million color images with varying sizes, belonging to $1,000$ classes. Each class has $1,300$ training images and $50$ validation images. 
ImageNet-100 is a subset of ImageNet1000 with 100 randomly selected classes, using the random seed of 1993.

\textbf{Protocols.} We adopt the protocol proposed in \citep{hou2019learning}, which is widely used in recent class-incremental learning works, to mimic realistic scenarios. The protocol involves an initial base task $\mathcal{T}_0$ and $T$ subsequent incremental tasks, where $T$ is set to $\{5,10,25,50\}$. For all datasets, we allocate half of the classes to the base task and distribute the rest equally among the incremental tasks. Moreover, we limit the number of exemplars per class to $20$ for reply.

\textbf{Metrics.} Our evaluation is based on the average incremental accuracy, as per \citep{rebuffi2017icarl}. This metric represents the mean of the accuracy of the models ${f^0, ..., f^T}$ acquired at each incremental step. Each model’s accuracy is calculated on all classes seen up to that step. In addition, we quantify forgetting by reporting the accuracy of the models ${f^0, ..., f^T}$ specifically on the initial base task $\mathcal{T}_0$.
\begin{figure*}[t!]
\centering
\begin{subfigure}{0.8\textwidth}
\includegraphics[width=\textwidth]{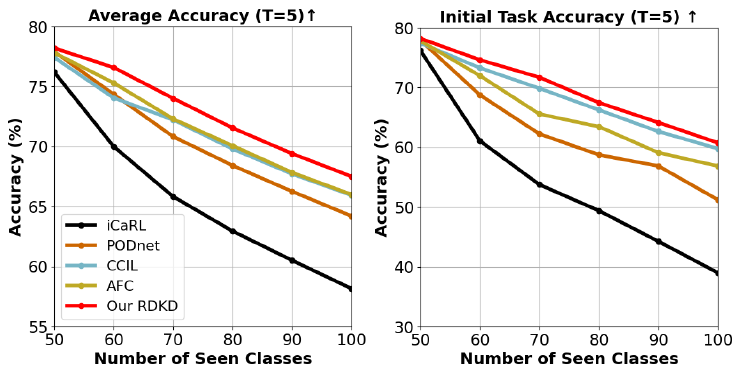}
\caption{\textbf{Results on CIFAR100 when T=5}}
\label{fig:flatness-a}    
\end{subfigure}
\begin{subfigure}{0.8\textwidth}
\includegraphics[width=\textwidth]{figures/fig-cifar5.pdf}
\caption{\textbf{Results on CIFAR100 when T=10}}
\label{fig:flatness-b}   
\end{subfigure}
\caption{\textbf{Results on CIFAR100.} Concerning both the average incremental accuracy and the accuracy of the initial task at each incremental step when $T=5$ and $T=10$, our method consistently outperforms throughout all stages. The minimal decline in accuracy on the initial task across incremental steps indicates reduced forgetting.}
\label{fig:cifar100} 
\end{figure*}

\textbf{Implementation details.} 
We base our experiments on the official code released by \cite{mittal2021essentials}. We use ResNet-18 for ImageNet100 and ImageNet1000, and ResNet-32 for CIFAR100. We train the network for 70 epochs with a learning rate of $1e-1$ for the base task, and for $40$ epochs with a learning rate of $1e-2$ for the incremental tasks. We reduce the learning rate by $10$ at epochs $\{30,60\}$ for the base task, and at epochs $\{25,35\}$ for the incremental tasks. For CIFAR100, we use a cosine learning rate schedule with a minimum of $1e-4$. We apply cosine normalization to the last layer, as recommended by \cite{hou2019learning}. We use SGD with a batch size of $128$, a momentum of $0.9$, and a weight decay of $1e-4$. $\lambda_{base}$ is set to $20$ for CIFAR100, $100$ for ImageNet100, and $600$ for ImageNet1000. We will share our codes publicly.

\textbf{Compared methods.}
We compare our method with six state-of-the-art class-incremental learning methods.
Among these, three utilize logit-based distillation: BIC \citep{wu2019large} and CCIL \citep{mittal2021essentials} employ GKD, while iCaRL \citep{rebuffi2017icarl} opts for TKD. The remaining three methods leverage feature-based distillation, including UCIR \citep{hou2019learning}, Podnet \citep{douillard2020podnet}, and AFC \citep{kang2022class}. Our experimental setup is consistent across all methods, and we either re-implement them or use their official code when available. In cases where some methods have reported results under identical settings, we incorporate their reported outcomes for comparison.

\begin{table*}[t!]
\centering
\footnotesize
\caption{\textbf{The effect of RDKD on ImageNet100.} RDKD exhibits the highest average incremental accuracy when compared across four distinct protocols. Furthermore, RDKD seamlessly integrates with established methods, leading to a notable improvement in accuracy across all protocols. The enhanced performance over existing methods is evidenced by the highlighted numbers in \textcolor{blue}{blue}.}
\resizebox{1.8\columnwidth}{!}{
\begin{tabular}{@{}lccccccccccccccccccccc@{}}
\toprule
&\multicolumn{4}{c}{\textbf{ImageNet100}}\\
\cmidrule(lr){2-5}
& \textbf{T=5} & \textbf{T=10} & \textbf{T=25} & \textbf{T=50}\\
\hline

BIC \citep{wu2019large} &68.97 &65.14 &59.65 &46.96 \\
iCaRL \citep{rebuffi2017icarl}&65.56 &60.90 &54.56 &54.97\\
UCIR \citep{hou2019learning} &71.04 &70.74 &62.94 &57.25 \\
Podnet \citep{douillard2020podnet} &75.54 &74.33 &68.31 &62.48 \\
CCIL \citep{mittal2021essentials} &77.83 &74.63 &67.41 &59.19 \\
AFC \citep{kang2022class} &76.65 &75.18 &72.91 &71.68 \\
\rowcolor{Gray}
RDKD (\textit{ours}) &\textbf{78.86}&\textbf{76.91} &73.16 &69.13 \\
\midrule
iCaRL+RDKD (\textit{ours}) &73.63(\textcolor{blue}{+8.07}) &69.95(\textcolor{blue}{+9.5}) &61.33(\textcolor{blue}{+6.77}) &57.54(\textcolor{blue}{+2.57})\\
CCIL+RDKD (\textit{ours}) &78.29(\textcolor{blue}{+0.46}) &76.77(\textcolor{blue}{+2.14}) &70.34(\textcolor{blue}{+2.93})  &65.82(\textcolor{blue}{+6.63})  \\
AFC+RDKD (\textit{ours}) &76.93 (\textcolor{blue}{+0.28}) &75.68(\textcolor{blue}{+0.49}) &\textbf{73.25}(\textcolor{blue}{+0.34}) &\textbf{72.34}(\textcolor{blue}{+0.66})  \\
\bottomrule
\end{tabular}}
\label{tab:imagenet100}
\end{table*}
\subsection{Main results}
\textbf{Results on CIFAR100}. We assess our method on CIFAR100 under four protocols: $T=5$, $T=10$, $T=25$, and $T=50$, corresponding to adding $10$, $5$, $2$, and $1$ classes per step, respectively. In Table~\ref{tab:cifar100}, our RDKD consistently outperforms other methods, particularly excelling at $T=5$ and $T=10$. For instance, RDKD boosts the average incremental accuracy from 66.49 to 67.51 for $T=5$ and from 64.41 to 65.06 for $T=10$, compared to the state-of-the-art AFC. However, AFC achieves superior results at $T=25$ and $T=50$, likely due to the challenge posed by a large number of tasks in the pool, where not all tasks have equal opportunity for selection during random distillation.

Furthermore, we integrate RDKD with existing methods, namely iCaRL, CCIL, Podnet, and AFC. For iCaRL and CCIL, we replace their TKD and GKD losses with our RDKD loss while maintaining other components unchanged. For Podnet and AFC, which utilize feature-based KD losses, we add our RDKD loss on top of their methods. Utilizing the official code of each method, we enhance the accuracy of all methods across all protocols, as shown in Table~\ref{tab:cifar100}. For instance, RDKD boosts AFC's accuracy from 66.49 to 67.13 at $T=5$.

In Fig. \ref{fig:cifar100}, we illustrate the average incremental accuracy and the initial task's accuracy for each step at $T=5$ and $T=10$. Our method consistently outperforms at all stages, demonstrating its efficacy in class-incremental learning. Notably, a key advantage of RDKD is its minimal degradation in accuracy on the initial task throughout incremental steps, indicative of reduced forgetting. This underscores our method's ability to retain previous knowledge while learning new tasks without compromising performance on any task.

\textbf{Results on ImageNet}
We show the comparison results on ImageNet100 with four protocols in Table \ref{tab:imagenet100}. Our method outperforms all the other methods on all protocols. When combined with existing methods, our RDKD improves the accuracy significantly. We also report the results on ImageNet1000 in Table \ref{tab:imagenet1000}. ImageNet1000 is the most difficult dataset, with more complex images than CIFAR100, and ten times more classes than ImageNet100. The results confirm the consistent advantage of our RDKD over the existing methods.
\begin{table*}[t!]
\centering
\footnotesize
\caption{\textbf{The effect of the proposed RDKD on ImageNet1000.} The proposed RDKD method demonstrates the highest average incremental accuracy among the compared approaches across two different protocols, including $T=5$ and $T=10$.}
\resizebox{1.98 \columnwidth}{!}{
\begin{tabular}{@{}lccccccccccccccccccccc@{}}
\toprule
& BIC \citep{wu2019large} & iCaRL \citep{rebuffi2017icarl}&UCIR \citep{hou2019learning}&Podnet \citep{douillard2020podnet}&AFC \citep{kang2022class}&RDKD (\textit{ours})\\
\hline
T=5 &45.72&51.36&64.34&66.95&68.90&\textbf{69.62}\\
T=10 &44.31&46.72&61.28&64.13&67.02&\textbf{67.58}\\
\bottomrule
\end{tabular}}
\label{tab:imagenet1000}
\end{table*}

\subsection{Comparison with various distillation methods.}
\label{sec:analysis}
We conduct a comprehensive comparison between the proposed RDKD, as defined in Eq. (\ref{eq:rdkd}), and three alternative distillation approaches to highlight its superiority in terms of \textit{accuracy}, \textit{stability}, \textit{plasticity}, and \textit{the flatness of minima}. The first method, Global Knowledge Distillation (GKD), adheres to the formulation in Eq. (\ref{eq:gkd}). The second method, Task-wise Knowledge Distillation (TKD), follows the formulation in Eq. (\ref{eq:tkd}). The third method is our Full Dense Knowledge Distillation (FDKD), defined in Eq. (\ref{eq:fdkd}). 
\begin{table}[t!]
\centering
\footnotesize
\caption{\textbf{Comparison with different distillation methods on CIFAR100}. We assess the performance of the proposed RDKD against three distinct distillation methods: GKD, TKD, and FDKD. RDKD demonstrates enhanced average incremental accuracy while maintaining training time comparable to the widely used GKD.}
\resizebox{0.98 \columnwidth}{!}{
\begin{tabular}{@{}lccccccccccccccccccccc@{}}
\toprule
& \multicolumn{2}{c}{\textbf{T=5}} & \multicolumn{2}{c}{\textbf{T=10}}\\
\cmidrule(lr){2-3} \cmidrule(lr){4-5}
& Accuracy & Training time & Accuracy & Training time \\
\hline

GKD  &65.73 & \textbf{1.9h} &64.41&\textbf{2.2h}\\
TKD &66.65& 2.1h &64.70& 2.4h\\
FDKD (\textit{Ours})  &\textbf{67.53}& 2.8h &\textbf{65.13}& 3.1h\\
RDKD (\textit{Ours}) &67.51& 2.0h &65.06& 2.3h\\
\bottomrule
\end{tabular}}
\vspace{-4mm}
\label{tab:RDKD}
\end{table}

\textbf{Accuracy.} These experiments are conducted on CIFAR100 with protocols $T=5$ and $T=10$. All experimental settings remain consistent, with different distillation losses defined by the respective approaches. The reported results in Table \ref{tab:RDKD} include average incremental accuracy and training time (hours). Key observations include: 1) Both FDKD and RDKD outperform GKD and TKD; 2) RDKD achieves a similar effect to FDKD with reduced training time.

\begin{figure*}[t!]
\centering
\begin{subfigure}{0.41\textwidth}
\includegraphics[width=\textwidth]{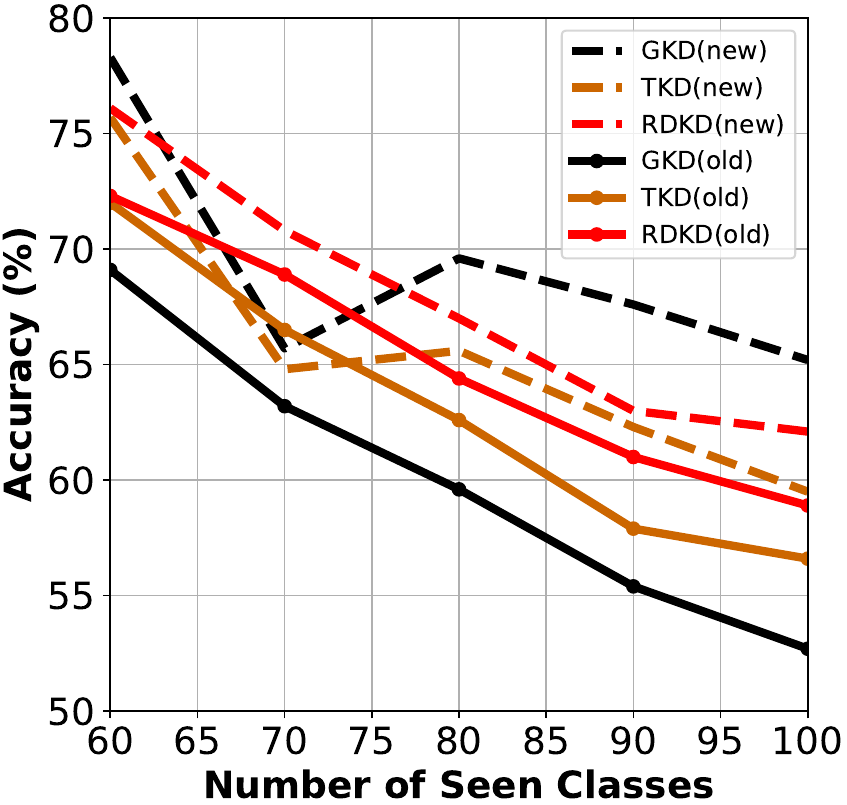}
\caption{\textbf{Accuracy on new and old classes}}
\label{fig:stability}    
\end{subfigure}
\begin{subfigure}{0.4\textwidth}
\includegraphics[width=\textwidth]{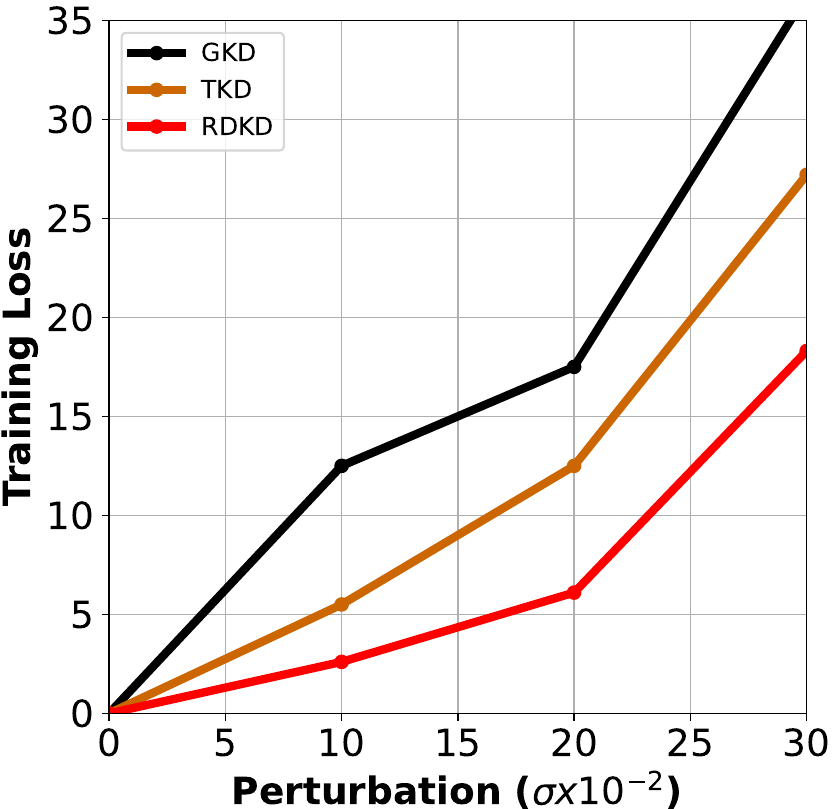}
\caption{\textbf{Sensitivity to perturbation}}
\label{fig:flatness}   
\end{subfigure}
\caption{\textbf{Enhancing stability and promoting flatter minima.} (a) In comparison to GKD and TKD, our RDKD consistently achieves higher accuracy on old classes across all stages. This indicates that RDKD reinforces stability more effectively than GKD and TKD; (b) RDKD exhibits lower sensitivity to perturbations compared to GKD and TKD, which highlights RDKD's capability to achieve flatter minima.}
\label{fig:analysis} 
\end{figure*}

\textbf{Stability and plasticity.}
We further investigate stability and plasticity through an experimental analysis of classifier bias, where we compute the classifier's performance on old and new classes separately. This experiment is conducted on CIFAR100 with $T=5$. The results are shown in Fig. \ref{fig:stability}. In comparison to GKD and TKD, our RDKD consistently achieves higher accuracy on old classes across all stages. This indicates that RDKD reinforces stability more effectively than GKD and TKD, thereby preserving previous knowledge more robustly. Additionally, our RDKD consistently outperforms TKD in accuracy on new classes at all stages, albeit exhibiting slightly lower accuracy on new classes compared to GKD in later stages. While our RDKD prioritizes model stability to mitigate forgetting, it slightly compromises model plasticity. Nevertheless, overall, our approach strikes a more favorable balance between stability and plasticity compared to GKD and TKD, as supported by the superior overall accuracy presented in Table \ref{tab:RDKD}.

\textbf{Flatter minima.}
Furthermore, following \citep{buzzega2020dark}, we analyze the flatness of the training minima of GKD, TKD, and RDKD by assessing their sensitivity to parameter perturbations. Specifically, we examine the model's response at the end of training by introducing independent Gaussian noise with increasing $\sigma$ to each parameter. This enables us to measure its impact on the average loss across all training examples. As shown in Fig. \ref{fig:flatness}, RDKD exhibits lower sensitivity to perturbations compared to GKD and TKD. This highlights RDKD's capability to achieve flatter minima, which translates to improved generalization to testing data.

\subsection{Analysis and ablation studies}
\label{sec:ablation}
\textbf{Ablation study.}
Our RDKD has two components: the random distillation loss and the adaptive weighting scheme. We perform an ablation study on CIFAR100 with the protocols of $T=5$ and $T=10$ to showcase the contribution of each component. The results are shown in Table \ref{tab:ablation}. We can observe that each component clearly matters for improving performance. 
\begin{table}[t!]
\centering
\footnotesize
\caption{\textbf{Ablation study on CIFAR100.} Our RDKD has two components: the random distillation loss and the adaptive weighting scheme. Both components clearly matter for improving average incremental accuracy.}
\resizebox{0.98 \columnwidth}{!}{
\begin{tabular}{@{}lccccccccccccccccccccc@{}}
\toprule
\multicolumn{2}{c}{\textbf{RDKD}}&\multicolumn{2}{c}{\textbf{CIFAR100}}&\multicolumn{2}{c}{\textbf{ImageNet100}}\\
\cmidrule(lr){1-2} \cmidrule(lr){3-4} \cmidrule(lr){5-6}
Random distillation &Adaptive weighting &\textbf{T=5} & \textbf{T=10} &\textbf{T=5} & \textbf{T=10}\\
\hline
& &65.73 &64.40&77.52&74.63\\
& \checkmark&66.65 &64.87&78.24&75.86\\
\checkmark& &67.51 &65.06&78.86&76.91\\
\checkmark&\checkmark&\textbf{67.60} &\textbf{65.42}&79.18&77.31\\
\bottomrule
\end{tabular}}
\label{tab:ablation}
\end{table}
\begin{figure*}[t!]
\centering
\begin{subfigure}{0.4\textwidth}
\includegraphics[width=\textwidth]{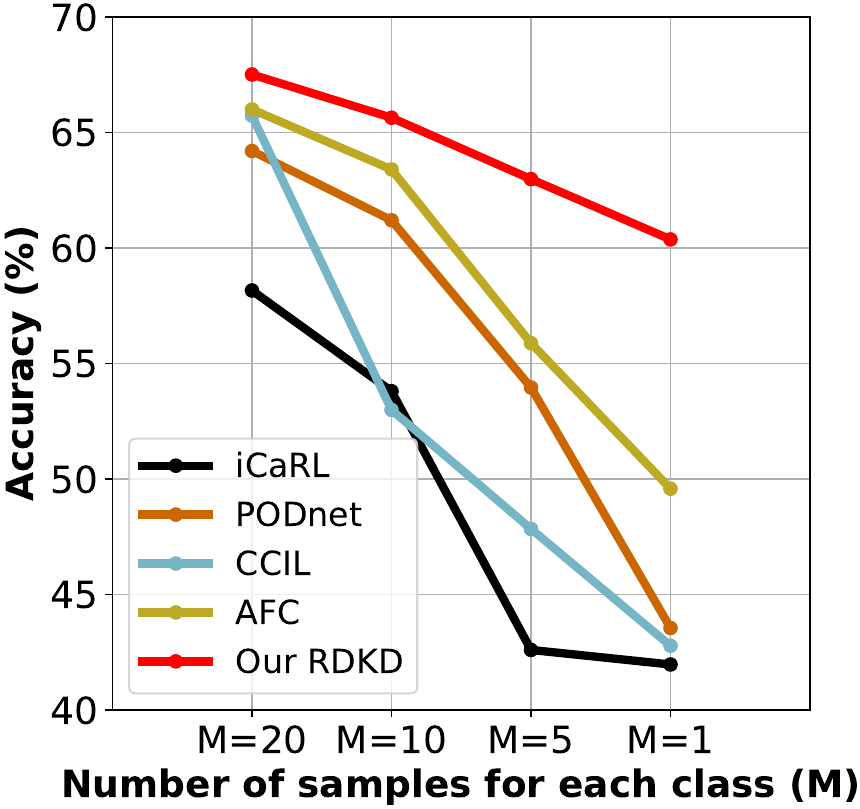}
\caption{\textbf{Memory sizes (T=5)}}
\label{fig:flatness-a}    
\end{subfigure}
\begin{subfigure}{0.4\textwidth}
\includegraphics[width=\textwidth]{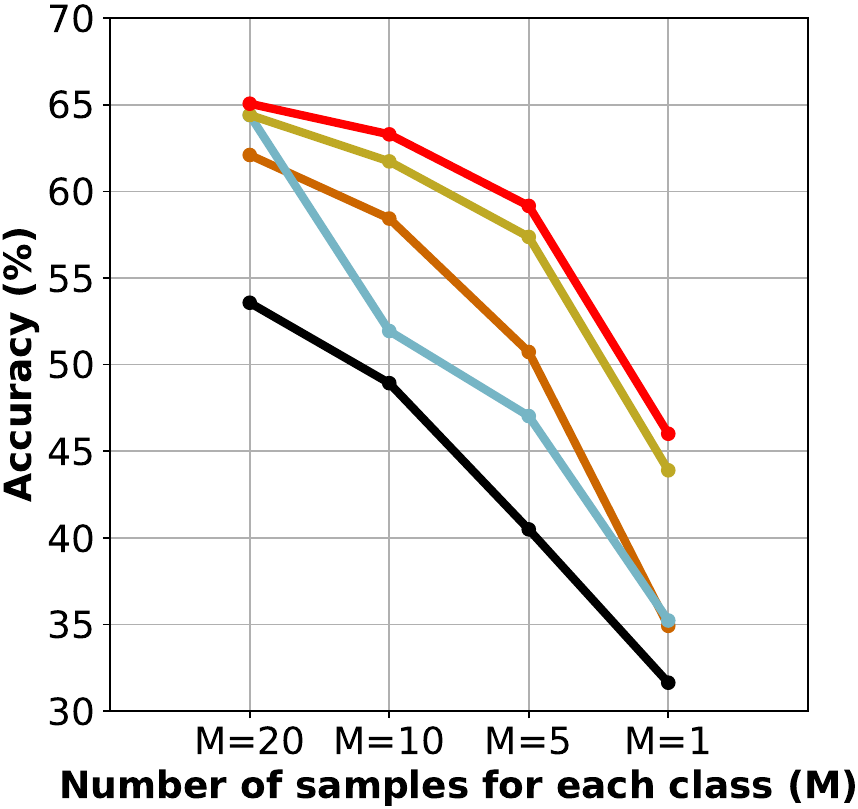}
\caption{\textbf{Memory sizes (T=10)}}
\label{fig:flatness-b}   
\end{subfigure}
\caption{\textbf{Robust to memory budget.} RDKD consistently surpasses other baselines across varying memory sizes in terms of the average incremental accuracy on CIFAR100. Despite a decrease in memory size, RDKD exhibits minimal accuracy degradation, showcasing robustness in exemplar memory size.}
\label{fig:memory} 
\end{figure*}

\textbf{Robust to memory budget.}
While prior results have already highlighted the significant improvement of RDKD over the baselines, we further emphasize its robustness to variations in memory budget. In Figure \ref{fig:memory}, we assess how RDKD compares against the baselines under different memory sizes (the number of samples stored for replay) per class, denoted as $M$. RDKD consistently outperforms other baselines across various scenarios, showcasing its robust applicability in diverse conditions. Notably, as the memory size decreases, the accuracy of all baselines experiences a sharp decline. In contrast, RDKD demonstrates the ability to maintain minimal accuracy degradation even with reduced memory size, highlighting its robustness concerning exemplar memory size.

\textbf{Robust to task orders.} 
Finally, we demonstrate that the proposed RDKD enhances the model's robustness to task order sensitivity, defined as the performance variation based on the task arrival sequence. We compare this to the baseline CCIL that utilizes GKD. Additionally, for a comprehensive assessment, we replace CCIL's GKD with our RDKD (CCIL+RDKD). The experimental results in Table \ref{tab:order} showcase the performance on order robustness for two models across five random sequences $\{S1, S2, S3, S4, S5\}$. RDKD consistently improves accuracy for all task orders compared to the baseline CCIL, reducing accuracy variance from 1.28 to 0.75. This reduction signifies that RDKD enhances the model's resilience to task order sensitivity.
\begin{table}[t!]
\centering
\footnotesize
\caption{\textbf{Robust to task orders}. Compared to the baseline CCIL, the proposed RDKD improves the model's robustness to the task order sensitivity.}
\resizebox{0.98 \columnwidth}{!}{
\begin{tabular}{@{}lccccccccccccccccccccc@{}}
\toprule
& S1 & S2 & S3 & S4& S5 & mean & variance \\
\hline
CCIL  &64.85 &67.3 &67.1 &67.44 &65.76 &66.49 &1.29\\
CCIL+RDKD (\textit{Ours}) &\textbf{66.09} &\textbf{67.95} &\textbf{67.61} &\textbf{68.27} &\textbf{66.99} &\textbf{67.38}&\textbf{0.75}\\
\bottomrule
\end{tabular}}
\label{tab:order}
\end{table}

\textbf{Effect on model compression.}
Finally, we showcase the versatility of our method in offline scenarios aimed at compressing a wide and deep teacher network into a shallower one. We conduct two experiments to demonstrate the superiority of RDKD over KD \citep{hinton2015distilling}. On CIFAR100, compressing a WideResNet-40-4 (teacher) to a ResNet-32 (student) using RDKD leads to an accuracy improvement from 75.16 (with KD) to 76.42. Similarly, on ImageNet100, compressing a WideResNet-28-10 to a ResNet-18 with RDKD results in an accuracy boost from 85.18 (with KD) to 87.32.

\section{Conclusion}
We introduce Dense Knowledge Distillation (DKD), a novel approach to mitigate forgetting in continual learning. DKD leverages a task pool to monitor the model's capabilities, dividing output logits into task-specific groups for distilling cumulative knowledge. To alleviate the computational burden, we incorporate random group selection. An adaptive weighting scheme ensures a balanced treatment of new and old classes based on their frequency and similarity. Across diverse benchmarks and class-incremental learning scenarios, our method outperforms state-of-the-art approaches. Empirical analysis underscores DKD's ability to enhance model stability, promote flatter minima for improved generalization, and maintain robustness under varying memory constraints and task orders. Furthermore, DKD seamlessly integrates with other continual learning methods to enhance their performance and demonstrates versatility in offline applications like model compression.


%
%

\bibliographystyle{spbasic}      
\bibliography{egbib}   

\end{document}